\documentclass[conference]{IEEEtran}
\IEEEoverridecommandlockouts

\usepackage{hyperref} 

\usepackage{multirow}
\usepackage{booktabs}
\usepackage{cite}
\usepackage{amsmath,amssymb,amsfonts}
\usepackage{algorithmic}
\usepackage{graphicx}
\usepackage{textcomp}
\usepackage{xcolor}
\usepackage{cleveref}
\crefname{figure}{fig.}{figures}
\Crefname{figure}{Fig.}{Figures}

\usepackage{caption}
\usepackage{graphicx}
\usepackage{float} 
\usepackage{subcaption}

\usepackage{graphicx}

\usepackage{cutwin}
\usepackage{enumitem}
\usepackage{amsmath}
\usepackage{booktabs}
\usepackage{multirow}
\usepackage{color}
\usepackage{colortbl}

\usepackage{amsfonts}
\usepackage{nicefrac}
\usepackage{microtype}
\usepackage{xcolor}
\usepackage{bbding}
\usepackage{pifont}
\usepackage{placeins}
\usepackage{subcaption} 

\definecolor{cvprblue}{rgb}{0.21,0.49,0.74}

\usepackage{booktabs} 
\usepackage{multirow} 
\usepackage{array}    
\usepackage{adjustbox} 
\def\BibTeX{{\rm B\kern-.05em{\sc i\kern-.025em b}\kern-.08em
    T\kern-.1667em\lower.7ex\hbox{E}\kern-.125emX}}

\usepackage{titlesec}  

\titlespacing*{\section}{0pt}{0.5ex}{0.5ex}
\titlespacing*{\subsection}{0pt}{0.5ex}{0.5ex} 

\usepackage{array}

\begin{document}

\title{TACR-YOLO: A Real-time Detection Framework for Abnormal Human Behaviors Enhanced with Coordinate and Task-Aware Representations
}

\author{
\IEEEauthorblockN{
Xinyi Yin\textsuperscript{1},
Wenbo Yuan\textsuperscript{1},
Xuecheng Wu\textsuperscript{2*}\thanks{*Corresponding author: wuxc3@stu.xjtu.edu.cn},
Liangyu Fu\textsuperscript{3},
Danlei Huang\textsuperscript{2}
}
\IEEEauthorblockA{
\textsuperscript{1}\textit{School of Cyber Science and Engineering, Zhengzhou University, Zhengzhou, Henan, China}
}
\IEEEauthorblockA{
\textsuperscript{2}\textit{School of Computer Science and Technology, Xi’an Jiaotong University, Xi’an, Shaanxi, China}
}
\IEEEauthorblockA{
\textsuperscript{3}\textit{School of Software, Northwestern Polytechnical University, Xi’an, Shaanxi, China}
}
\IEEEauthorblockA{
Emails: yinxinyi@stu.zzu.edu.cn, rose2835228047@gmail.com,\\
wuxc3@stu.xjtu.edu.cn, lyfu@mail.nwpu.edu.cn, forsummer@stu.xjtu.edu.cn
}
}

\maketitle

\begin{abstract}
Abnormal human behavior Detection (AHBD) under special scenarios is becoming increasingly crucial. While YOLO-based detection methods excel in real-time tasks, they remain hindered by challenges including small objects, task conflicts, and multi-scale fusion in AHBD. To tackle them, we propose TACR-YOLO, a new real-time framework for AHBD. We introduce a Coordinate Attention Module to enhance small object detection, a Task-Aware Attention Module to deal with classification-regression conflicts, and a Strengthen Neck Network for refined multi-scale fusion, respectively. In addition, we optimize Anchor Box sizes using K-means clustering and deploy DIoU-Loss to improve bounding box regression. The \textit{\underline{P}}ersonnel \textit{\underline{A}}nomalous \textit{\underline{B}}ehavior \textit{\underline{D}}etection (PABD) dataset, which includes 8,529 samples across four behavior categories, is also presented. Extensive experimental results indicate that TACR-YOLO achieves 91.92\% mAP on PABD, with competitive speed and robustness. Ablation studies highlight the contribution of each improvement. This work provides new insights for abnormal behavior detection under special scenarios, advancing its progress.
\end{abstract}


\section{Introduction}

As society advances, industries face growing challenges in managing personnel behavior. Actions like using mobile phones or smoking during construction can disrupt operations, reduce work quality, and increase accident risks\cite{1}. Therefore, accurately and promptly analyzing abnormal behavior in specific scenarios holds substantial practical value.

Recent progress in deep learning—especially in real-time object detection powered by convolutional neural networks (CNNs)—has greatly improved the effectiveness of abnormal behavior analysis. Object detection techniques are generally categorized into two-stage and one-stage approaches\cite{2}. Two-stage models, such as R-CNN\cite{3}, Fast R-CNN\cite{4}, and Faster R-CNN\cite{faster}, are known for their high accuracy but suffer from heavy computational demands, making them less suitable for real-time scenarios\cite{2, 25}. In contrast, one-stage detectors—like the YOLO family\cite{yolo, yolov3, yolov4, yolov6, yolov7, yolov8}, SSD\cite{ssd}, Retina-Net\cite{Retina-Net}, and CenterNet++\cite{CenterNet++}—frame detection as a regression problem, achieving an optimized equilibrium between computational efficiency and detection accuracy. Among them, the YOLO series\cite{25} stands out for its innovative design, simplified training process, and achieve an optimal balance between computational efficiency and detection accuracy, making it a benchmark for real-time detection\cite{5}. However, existing YOLO algorithms still face some challenges when addressing abnormal behavior detection tasks in special scenarios: (\textbf{i}) The single-stage detection structure of YOLO has limited capability to extract shallow features, resulting in weak perception of small objects\cite{2}. (\textbf{ii}) Due to the shared parameters for classification and regression tasks~\cite{30}, there is a potential conflict between the two, which becomes particularly significant in small object detection and complex scenarios. (\textbf{iii}) While a single feature fusion mechanism simplifies computation, it is insufficient to fully capture multiscale features in complex scenarios, which undermines performance, generalization, and robustness~\cite{26}.

To this end, based on YOLOv7-X~\cite{yolov7}, we proposed an improved real-time framework denoted TACR-YOLO, suitable for AHBD in special scenarios. First of all, to enhance small object detection (\textit{e.g.}, cigarette butts, hands), we integrate a Coordinate Attention Module into the middle network, which decouples channel and spatial attention, improving sensitivity to small objects while expanding the receptive field to optimize large object localization. Furthermore, in response to the inherent divergence between classification and regression objectives, we have presented the Task-Aware Attention Module, which dynamically adjusts the feature weight distribution, enhancing the extraction of discriminative features without decoupling the classification and regression tasks. It effectively alleviates task inconsistency and feature coupling while maintaining minimal computational overhead. Finally, we introduce the Strengthen Neck Network to enhance multi-scale feature fusion, calibrate anchor dimensional parameters through K-means clustering to boost scale-aware detection efficacy, and implement the DIoU metric-driven localization optimization mechanism for improved regression precision, significantly improving detection performance and training stability.

Meanwhile, as existing datasets for object detection fail to meet the requirements for AHBD under special scenarios, we constructed a diverse dataset, PABD (\textbf{\textit{\underline{P}}}ersonnel \textbf{\textit{\underline{A}}}bnormal \textbf{\textit{\underline{B}}}ehavior \textbf{\textit{\underline{D}}}ataset), containing 8,529 images from scenarios like driving and construction sites. The dataset contains four label categories: phone, smoke, drink, and face (as shown in Fig. \ref{Dataset}). Data cleaning and augmentation techniques were applied to improve data balance, robustness, and diversity.

Experimental validation on the PABD dataset demonstrates TACR-YOLO's superior efficacy, achieving 91.92\% mAP with real-time inference speeds. Ablation analyses quantitatively substantiate each module's performance gains. Our principal contributions include:
\vspace{-0.77em}
\begin{enumerate}
\item We propose the TACR-YOLO framework and design a Task-Aware Attention Module to alleviate task inconsistency and feature coupling, significantly improving model performance and generalization.

\item By integrating the Coordination Attention Module into YOLOv7-X's backbone network, optimizing anchor box sizes with K-means clustering, designing the Strengthen Neck Network, and introducing DIoU-Loss, we enhance the detection capability through a multi-perspective hierarchical approach.

\item We construct the PABD dataset, tailored for this task, covering diverse scenarios and addressing the lack of data in this field. Comprehensive evaluation on the PABD dataset validates the effectiveness of our novel framework.
\end{enumerate}

\section{Related Work}

\begin{figure*}[t!]
\centering
\includegraphics[width=\linewidth]{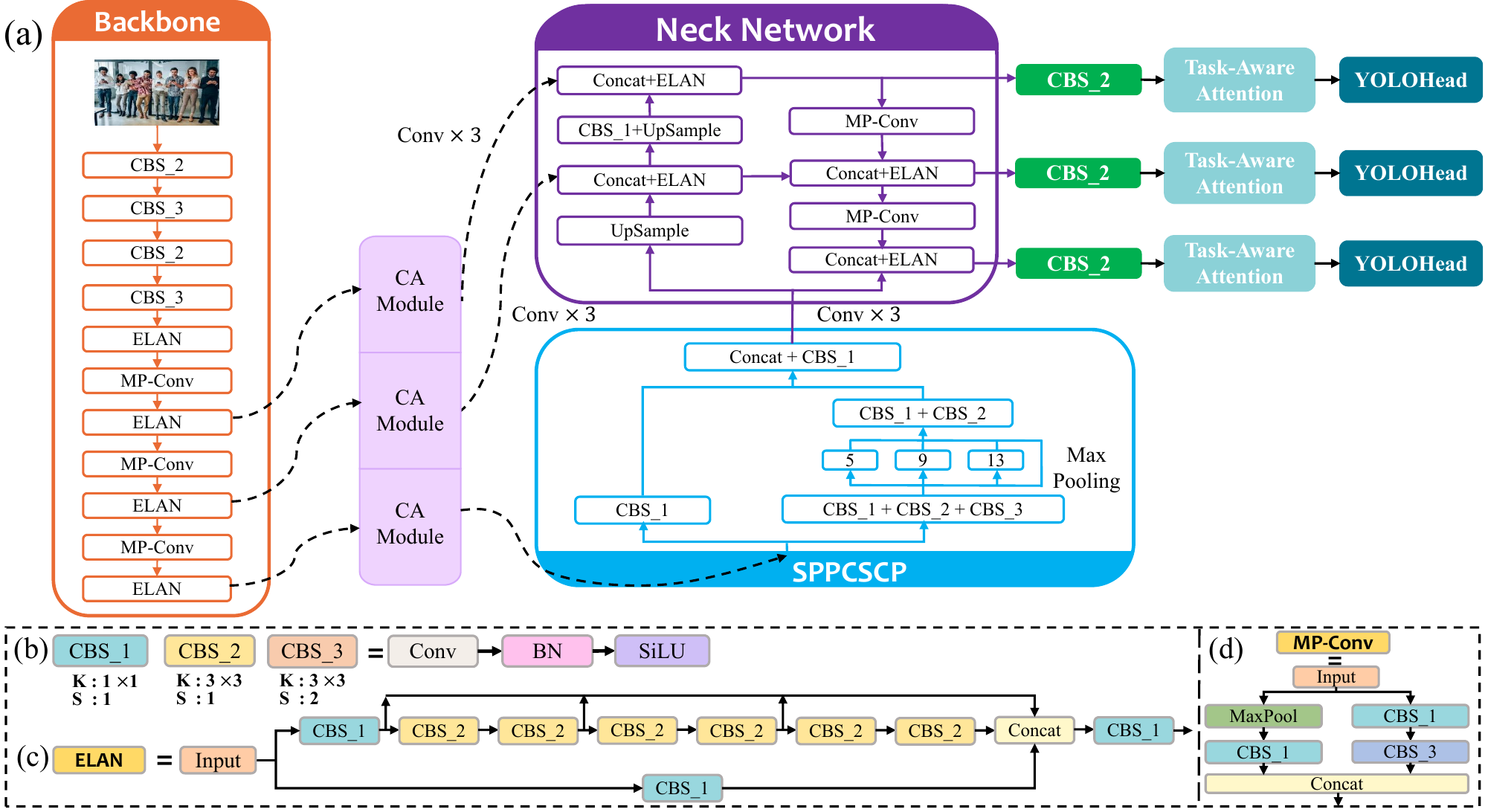}
\caption{(a) illustrates the overall TACR network structure, where SPPCSCP denotes multi-scale feature fusion. (b) details the Normalized Swish Convolution Block, where $K$ is the kernel size and $S$ is the stride. (c) displays the ELAN (Efficient Layer Attention Network) structure, where \textit{Concat} refers to feature concatenation. (d) depicts MP-Conv (Multi-path Convolution), with MaxPool representing max pooling.}
\label{overall framework}
\vspace{-2em}
\end{figure*}

\subsection{Advancements in Object Detection Mthods}

The field of object detection has undergone revolutionary transformation through innovations in deep learning, with convolutional neural network (CNN) approaches systematically classified into two-stage paradigms and one-stage architectures \cite{2}. Two-stage methods \cite{3, 4, faster} involve generating candidate regions followed by classification and localization. R-CNN \cite{3} used selective search for region proposals and CNNs for feature extraction but suffered from slow speeds due to fixed input sizes \cite{36}. SPP-Net \cite{35} eliminated input scaling but required separate fine-tuning and regression steps, adding complexity \cite{37}. Fast R-CNN \cite{4} introduced end-to-end training but still relied on selective search, limiting efficiency \cite{36}. Faster R-CNN \cite{faster} improved efficiency by introducing the Region Proposal Network (RPN), which shared convolutional features and reduced computational cost. However, RPN generates numerous proposals and classifies them independently, slowing inference and limiting real-time performance \cite{37}.

In contrast, one-stage object detection approaches \cite{yolo, yolov3, yolov4, yolov6, yolov7, yolov8, ssd, Retina-Net, CenterNet++} simplify detection into a unified regression problem, achieving simultaneous category prediction and bounding box estimation through a single convolutional network \cite{36}. While slightly less accurate than two-stage methods, they offer a significant speed advantage. Among them, the YOLO series excels in real-time performance and detection accuracy, with YOLOv7-X \cite{yolov7} further enhancing multi-scale adaptability and inference efficiency through an optimized structure and improved task assignment. Its balance of speed and accuracy makes YOLOv7-X ideal for real-time tasks, and we selected it as our baseline model, refining it for faster and more accurate detection in real-time scenarios.

\subsection{Abnormal Human Behavior Detection}

Detecting abnormal behavior has become a key research area in object detection, with many advancements in recent years. One-stage detection algorithms, particularly the YOLO series, are favored for their speed \cite{31}. For instance, \cite{40} introduced YOLOv3-Multi for pedestrian detection, improving small object detection with a residual DarkNet structure and spatial pyramid pooling (SPP). \cite{38} proposed a lightweight MobileNetv3-based YOLOv3, replacing DarkNet53 to reduce complexity, and added CIoU and SESAM attention for better long-distance detection. RSA-YOLO \cite{33} addressed aspect ratio and small object detection but faces high computational complexity in large-scale images, slowing inference speeds. Other approaches combine YOLO with CNNs, such as \cite{39}, which uses YOLO with 2D convolutional layers for real-time abnormal human behavior detection in videos. \cite{32} applied YOLOv5 to identify abnormal behaviors in videos, using CNNs for motion feature extraction. Although effective for real-time detection, these methods still struggle with small objects and complex scenes.

In summary, YOLO-based methods for abnormal behavior detection excel in real-time detection and small object recognition but face challenges in inference speed and performance in complex scenarios. Additionally, few studies address the conflicts between classification and regression tasks that impact performance.

\section{Methodology}

\subsection{TACR-YOLO Overview}

To achieve faster and more accurate abnormal behavior detection, we propose TACR-YOLO based on YOLOv7-X, as shown in Fig. \ref{overall framework}. The network consists of four main components: the input module, backbone feature extraction, enhanced feature extraction, and output module.

First, in the input module, images are preprocessed by resizing to 640×640 pixels and normalizing to RGB format. The K-means algorithm is applied to cluster bounding box dimensions in the training set, generating dataset-specific anchor boxes. This improves the model's generalization, detection accuracy, and robustness across varying target sizes.

The backbone feature extraction network is built on YOLOv7-X’s MP-Conv and ELAN structures, with enhancements to feature connections and gradient flow paths, improving feature representation and computational efficiency. The backbone outputs feature maps at three scales (feat1: 80 $\times$ 80 $\times$ 512, feat2: 40 $\times$ 40 $\times$ 1024, feat3: 20 $\times$ 20 $\times$ 1024), which are then passed through the Coordinate Attention module. This module encodes spatial positional information and generates channel attention weights, boosting the network's focus on key regions and improving small object detection (\textit{e.g.}, mobile phones, cigarette butts). The design of the Coordinate Attention module not only considers inter-channel dependencies but also fully utilizes spatial information, helping the model precisely locate and focus on critical areas, thereby improving detection performance.

In the neck network, we propose an enhanced neck structure designed for upsampling, feature integration, and channel adjustment. This structure adopts a multi-branch feature fusion strategy, where features from different branches are processed via convolution and then stacked together. Features from multiple branches are first processed through convolutional layers and then fused by stacking, following a multi-branch feature fusion strategy adopted by this structure. This effectively captures and integrates multi-scale features, generating feature layers with enriched semantic information while maintaining the same size.

Finally, the YOLO head incorporates the Task-Aware Attention module, thereby boosting the model's capacity to manage multiple detection tasks effectively. Together, these improvements allow TACR-YOLO to achieve outstanding performance on the PABD dataset.


\subsection{Coordinate Attention Module}

By encoding precise positional cues into channel-wise feature modulation, the Coordinate Attention (CA) module~\cite{41} facilitates more effective multi-scale object detection through improved contextual sensitivity. In the shallow feature layer (feat1), it enhances the capture of fine details in small targets (e.g., phones, cigarette butts), ensuring accurate location and boundary extraction. In the intermediate feature layer (feat2), the CA module optimizes multi-scale feature fusion and robustness for medium-scale targets (e.g., cups). In the deep feature layer (feat3), it strengthens deep semantic feature extraction, improving the detection of high-level target characteristics. By applying CA at all the stages of feature extraction, the model leverages spatial and semantic information across scales, enhancing detection accuracy and robustness. The module operates via dual-phase processing: (i) positional embedding, and (ii) spatial attention formulation.

(i)~\textbf{Positional Embedding:} For a given input $X$, directional pooling is performed using kernels of size $(H,1)$ and $(1,W)$ to capture channel-wise features along vertical and horizontal axes, respectively. This process aggregates features across these dimensions, effectively converting conventional 2D global pooling into compact 1D feature representations. This enables the CA module to capture precise spatial interactions, even with remote spatial context. As shown in Equations (1) and (2) below, two direction-sensitive feature maps, $z_c^h (h)$ and $z_c^w (w)$ are generated, which can facilitate the extraction of fine-grained positional details and enhance the captures of remote spatial interactions, \textit{i.e.},
\vspace{0.5em}
\begin{align}
z_c^h (h) &= \frac{1}{W}\displaystyle\sum_{w=1}^{W}{x}_{c}(h, w),\\
z_c^w (w) &= \frac{1}{H}\displaystyle\sum_{h=1}^{H}{x}_{c}(h, w),
\end{align}
where H and W represent the height and width of the feature map, respectively.

(ii)~\textbf{Spatial Attention Formulation: }The direction-aware feature maps are initially concatenated and passed through a 1 $\times$ 1 convolution to reduce the channel dimension to $C/r$. Batch Normalization (BatchNorm) and ReLU are then employed to refine the spatial encoding of vertical and horizontal positional features, \textit{i.e.},
\begin{equation}
f = \delta ({F}_{1}[{Z}^{h}, {Z}^{w}]),
\vspace{-0.5em}
\end{equation}
where ${F}_{1}(\cdot )$ corresponds to a 1$\times$1 convolutional layer, $\delta(\cdot )$  indicates the ReLU nonlinearity, and $f\in{\mathbb{R}}^{\frac{C}{r}\times(H+W)}$.

Then, the feature map $f$ undergoes spatial decomposition along orthogonal axes, generating decoupled feature representations ${f}^{h}$ and ${f}^{w}$. These are then undergo independent convolutional and sigmoidal transformations to yield axis-specific attention weights ${g}^{h}$ and ${g}^{w}$, as defined in Equations (4) and (5),

\vspace{-1.0em}
\begin{align}
{g}^{h} &= \sigma ({F}_{h}({f}^{h})),\\
{g}^{w} &= \sigma ({F}_{w}({f}^{w})),
\vspace{0.5em}
\end{align}
where ${F}_{h}(\cdot)$ and ${F}_{w}(\cdot)$ represent the Conv2D operations, and $\sigma (\cdot)$ denotes the Sigmoid activation function.

Finally, ${g}^{h}$ and ${g}^{w}$ are extended and applied as coordinate attention weights, culminating in the module's final output formulation expressed in Equation (6), \textit{i.e.},

\vspace{-0.6em}
\begin{equation}
{y}_{c}(i, j) = {x}_{c}(i, j) \times {g}_{c}^{h}(i) \times {g}_{c}^{w}(j),
\vspace{-0.4em}
\end{equation}
where, ${x}{c}(i, j)$ is the input feature, ${g}{c}^{h}(i)$ and ${g}{c}^{w}(j)$ are vertical and horizontal weights, and ${y}{c}(i, j)$ is the output after applying these weights.

\subsection{Task-Aware Attention Module}
Single-stage detectors perform integrated prediction tasks within a unified framework, including localization, category recognition, and confidence estimation. However, the coupling of feature distributions between regression and classification tasks often leads to poor performance in object localization and classification, especially for small- to medium-scale targets in complex scenarios like phones, cigarette butts, and cups. To address this, we designed the Task-Aware Attention Module based on DY-ReLU-A\cite{42}, aiming to better express and generalize the relationships between tasks.

\textbf{DY-ReLU-A (Dynamic ReLU with Attention)} is a dynamic activation function that adjusts its threshold to optimize feature responses based on task requirements, addressing conflicts between classification and regression tasks. It uses a global information encoding module to capture contextual features, such as target size, location, and class distribution, via global average pooling. In classification tasks, DY-ReLU-A enhances feature discriminability, while in regression tasks, it significantly improves target localization accuracy. Additionally, it refines the model’s response to diverse object sizes and background complexities through adaptive channel modulation.

In the \textbf{Task-Aware Attention Module}, DY-ReLU-A is the core component, dynamically adjusting the activation function to optimize feature representations for different tasks. The module first receives feature tensors from the feature extraction network (e.g., FPN), which contain multi-level object information and spatial context. These multi-task features are integrated through the Concat Layer to match the input format for the Task-Aware Layer. DY-ReLU-A is introduced as the activation function, using a global information encoding module to extract global contexts, such as class distribution, spatial relationships, and scale variations. Spatial information is first compressed into a global feature vector via global average pooling, followed by two fully connected layers with normalization to produce dynamic parameters. These parameters control the activation function, allowing it to adjust feature response intensity for classification and regression tasks. The dynamically generated weights are applied to 
the activation function, improving the handling of feature coupling. The process is formulated as,
\begin{equation}
{\pi }_{c}(F)×F = \max_{}{({\alpha }^{1}(F){F}_{c} + {\beta }^{1}(F), {\alpha }^{2},(F)×{F}_{c} + {\beta }^{2}(F))},
\end{equation}
where ${F}_{c}$ denotes the channel-wise activation subset within the $c$-th feature map, with ${\alpha }^{1}, {\beta }^{1}, {\alpha }^{2} and {\beta }^{2}$ constituting the learnable modulation coefficients of the dynamic ReLU operator.

In our experiments, the incorporation of the Task-Aware Attention Module significantly enhances the model's mAP in complex scenarios, validating its powerful capabilities in task awareness and feature optimization.

\subsection{Strengthen Neck Network}
In the original YOLOv7-X\cite{yolov7}, the enhanced feature extraction network improves deep feature representation and feature fusion. However, the network faces challenges when dealing with small objects, low-texture regions, and cluttered scenes. Limited convolutional depth may hinder the extraction of detailed spatial and semantic features, especially for small-scale targets, thereby compromising both classification precision and localization accuracy in multi-scale detection tasks.

To counter these shortcomings, we proposed architectural refinements to the enhanced feature extraction network in YOLOv7-X by replacing the single convolution operation before feeding feat1, feat2, and feat3 into the neck network with three convolution operations. This improvement augments the hierarchical depth of convolutional operations, enabling the network to extract multi-scale features of targets more comprehensively at different levels while enhancing its capacity to resolve fine-grained details in small object instances.

The experimental results show that this modification enhances the network's depth and capacity, effectively addressing small target challenges without increasing computational cost. It also improves robustness and generalization, offering a more reliable solution.

\subsection{Other tricks for enhancement }
\noindent \textbf{(i) Adaptive Anchor Box Design}: During dataset pre-processing, the K-means algorithm clusters bounding box dimensions in the training dataset to optimize the specific anchor box sizes. By minimizing distances to centroids, it identifies box dimensions that enhance detection accuracy and generalization. Based on common practice in object detection models such as YOLO, the number of clusters K is set to 9, as inspired by the COCO dataset. This process automatically adjusts Anchor Boxes to target size and shape variations, as shown in Equation (8) below:
\begin{equation}
J = \displaystyle\sum_{i = 1}^{K}\displaystyle\sum_{j=1}^{n}{\parallel {x}_{j}-{\mu }_{i}\parallel}^{2},
\end{equation}
where $J$ denotes the sum of squared intra-cluster errors, $K$ specifies the cluster count, $n$ indicates the total data points, ${x}_{j}$ corresponds to the $j$-th sample, ${\mu }_{i}$ defines the centroid of the $i$-th cluster, and ${\parallel {x}_{j}-{\mu }_{i}\parallel}^{2}$ denotes the squared Euclidean distance between ${x}_{j}$ and ${\mu }_{i}$.

\noindent \textbf{(ii) Loss Function Optimization:} The Intersection over Union (IoU) metric, while widely adopted in object detection tasks to quantify bounding box overlap, presents notable drawbacks. A critical limitation arises when predicted and ground truth boxes exhibit zero overlap: The IoU value collapses to zero regardless of their relative positions. What's more, when two boxes share the same size and IoU, IoU Loss cannot capture their positional differences, reducing accuracy, especially in cases where the box positions vary significantly.

DIoU Loss~\cite{DIOU} improves upon IoU-based losses by incorporating a center-distance penalty between predicted and ground truth boxes, thus improving localization accuracy through better spatial alignment. Doing this improves optimization, particularly for boxes with similar positions but the same size, and resolves IoU's non-differentiability. DIoU also enhances small object detection (e.g., phones, cigarette butts) by improving localization and mitigating gradient vanishing, boosting overall performance. The DIoU calculations are shown as follows:

\begin{align}
IOU  &= \frac{A\cap B}{A\cup B}, \\
DIOU &= IOU - \frac{{\rho }^{2}(b, {b}^{gt})}{{c}^{2}},
\end{align}
where $b$ and ${b}^{gt}$ refer to the center points of the predicted and ground truth boxes, $\rho$ represents the Euclidean distance between them, and $c$ denotes the diagonal length of their smallest enclosing rectangle.

\section{PABD Dataset}

\begin{figure}[t!]
\centering
\includegraphics[width= \linewidth, height=0.23\textheight]{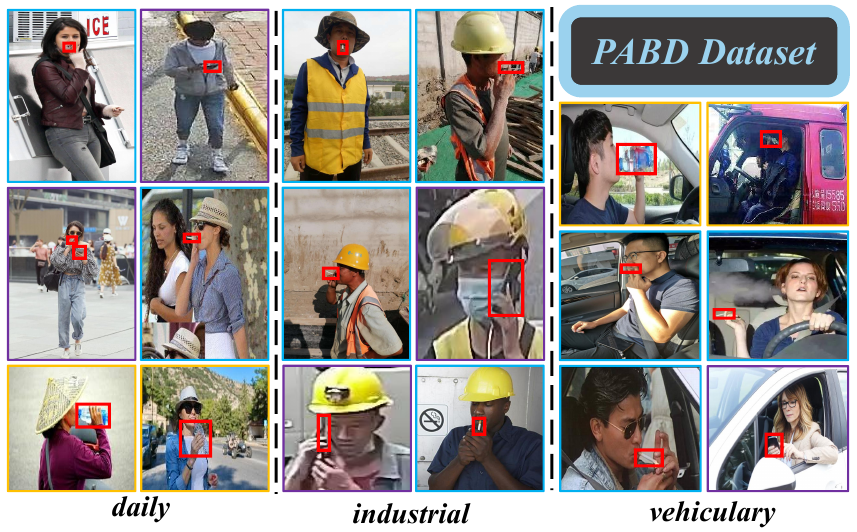} 
\caption{The PABD dataset consists of 8,529 images across three main scenarios, with frame border colors indicating behavior categories: \textcolor[rgb]{0, 0.69, 0.941}{smoke}, \textcolor[rgb]{0.439, 0.188, 0.627}{phone}, and \textcolor[rgb]{1.0, 0.8039, 0.2}{drink}.}
\label{Dataset}
\vspace{-0.7em}
\end{figure}

Currently, datasets for detecting abnormal personnel behaviors in natural scenes are relatively scarce. As a result, the Personnel Anomaly Behaviors Detection (PABD) dataset was constructed, as shown in Fig.~\ref{Dataset}. It contains a total of 8,529 images captured in various typical public natural settings, such as subway stations and shopping malls. The dataset demonstrates a high degree of diversity and broad adaptability, supporting the application needs of models across a range of complex scenarios. Table~\ref{tab:dataset-stats} summarizes the statistical characteristics of our PABD dataset.

The dataset construction involved three stages: (1)~\textbf{Collection}: images were sourced from public platforms and web crawlers targeting specific scenarios (\textit{e.g.}, subway stations, malls) for diversity. 
(2)~\textbf{Filtering:} low-quality, blurred, and redundant samples were removed through automated and manual review.
(3)~\textbf{Annotation: }labelImg was used for annotation with Pascal VOC format bounding boxes, followed by multiple verification rounds for consistency.

Data cleaning (\textit{e.g.}, balancing categories, removing duplicates) and augmentation (\textit{e.g.}, geometric transformations, color adjustments, noise addition) enhanced dataset diversity and model robustness.

Ultimately, the dataset is split into training, validation, and test sets to support effective model training and evaluation. Benefiting from this rigorous data processing pipeline, the PABD dataset excels in sample scale, class balance, data quality, and adaptability, enabling its effective application in complex scenarios involving abnormal behavior recognition and real-time tracking.

\section{Experiments}

\subsection{Evaluation Metrics}

This investigation employs two principal evaluation metrics: Average Precision (AP) and mean Average Precision (mAP). The AP metric operates within a holistic evaluation framework incorporating precision (P) and recall (R), mathematically represented as:

\vspace{-1.0em}
\begin{equation}
P = \dfrac{TP}{TP+FP},
\end{equation}
\begin{equation}
R = \dfrac{TP}{TP+FN},
\vspace{-0.5em}
\end{equation}
where $TP$, $FP$, and $FN$ denote the counts of true positives, false positives, and false negatives, respectively.

\begin{table}[t!]
\centering
\caption{The statistics and distributions of PABD dataset.}
\begin{tabular}{>{\centering\arraybackslash}p{0.8cm}|>{\centering\arraybackslash}p{0.4cm}>{\centering\arraybackslash}p{0.4cm}>{\centering\arraybackslash}p{0.4cm}>{\centering\arraybackslash}p{0.4cm}|>{\centering\arraybackslash}p{0.6cm}>{\centering\arraybackslash}p{0.6cm}>{\centering\arraybackslash}p{0.6cm}>{\centering\arraybackslash}p{0.6cm}}
\hline
Dataset & Total  & Train & Val & Test & drink & face & phone & smoke \\ \hline
PABD & \textbf{8529}  & 6908 & 768 & 853 & 1128 & 7959 & 3413 & 3300 \\ \hline
\end{tabular}
\label{tab:dataset-stats}
\vspace{-0.5em}
\end{table}

The average precision (AP) evaluates detection efficacy by calculating the integral under the precision-recall (PR) curve. The mAP, as the mean of APs across all categories, offers a comprehensive measure of overall model accuracy.
\vspace{0.5em}
\begin{align}
AP  &= \int PR dR, \\
mAP &= \dfrac{\sum_{i}^{c} C_i}{c},
\end{align}
where $C_i$ denotes the AP metric for individual object categories and $c$ specifies the total category count within the detection framework.

\subsection{Implementation Details}

To determine the most suitable hyperparameters for the baseline model, We performed hyperparameter tuning experiments for YOLOv7-X on the PABD dataset.

\begin{figure*}[t!]
\centering
\includegraphics[width=0.90\linewidth, height=8cm, keepaspectratio=false]{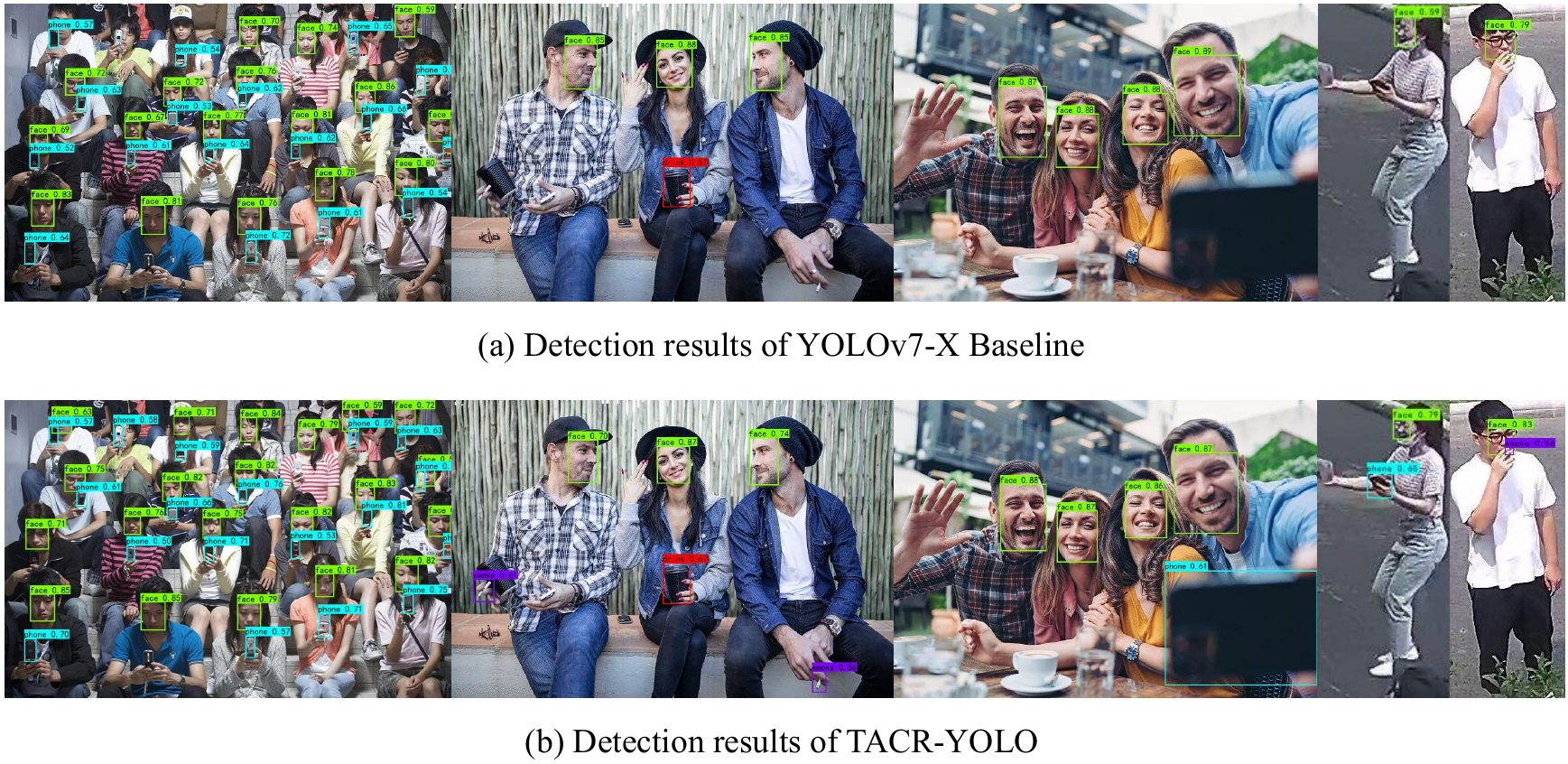}
\caption{The detection results of the YOLOv7-X Baseline and our proposed TACR-YOLO.}
\vspace{-1.5em}
\label{detection_comparsion}
\end{figure*}
In this study, we train TACR-YOLO in an end-to-end manner for 100 epochs using input images resized to 640$\times$640. The training is conducted using two NVIDIA V100 GPUs. Through this series of experiments, we finally choose to apply the Cosine Annealing LR and turn on mosaic data augmentation. During training, the model undergoes 30 epochs of frozen training with a batch size of 8, followed by 70 epochs of unfrozen training using a batch size of 4. Optimization is carried out with the SGD optimizer, employing a base learning rate of 0.01 and a weight decay of 5\textit{e}-4. Through these hyperparameter tuning experiments, we achieved the best performance with YOLOv7-X, obtaining an mAP of 88.68\%. Table~\ref{hyperparameter-tunning} presents the results of the hyperparameter tuning experiments.

\begin{table}[t!]
\centering
\caption{The Hyper-parameter adjustment studies of our proposed method.}
\resizebox{\linewidth}{!}{
\begin{tabular}{
            >{\centering\arraybackslash}m{2.65cm} 
            *{4}{>{\centering\arraybackslash}m{0.72cm}} 
            >{\centering\arraybackslash}m{0.78cm} 
        }
\toprule
\multirow{2}{*}{Methods} & \multicolumn{4}{c}{AP (\%)} & \multirow{2}{*}{mAP (\%)} \\
            \cmidrule(lr){2-5}
            & drink & face & phone & smoke & \\
            \midrule
            
            Baseline+StepLR+no mosaic+SGD+30\textbar100 & 90.26 & 91.04 & 79.67 & 84.43 & 86.35\\
            Baseline+Cosine+no mosaic+SGD+30\textbar100 & 93.89 & 93.42 & 81.98 & 84.75 & 88.51\\
            \textbf{Baseline+Cosine+
            mosaic+SGD+30\textbar100} & 93.71 & 93.01 & 81.16 & 86.82 & \textbf{88.68}\\
            Baseline+Cosine+
            mosaic+Adam+30\textbar100 & 93.33 & 92.99 & 81.32 & 84.56 & 88.05\\
            Baseline+Cosine+
            mosaic+SGD+40\textbar120 & 94.29 & 92.88 & 81.71 & 85.00 & 88.47\\
            \bottomrule
        \end{tabular}}
\label{hyperparameter-tunning}
\end{table}

\begin{table}[t!]
\centering
\caption{Performance comparisons of TACR-YOLO and other advanced real-time object detection methods on the PABD dataset. Note that we highlight the best performance in \textbf{\textit{bold}} and \underline{underline} the second performance.}
\resizebox{\linewidth}{!}{
   \begin{tabular}{lcccccc} 
        \toprule
        \multirow{2}{*}{Methods} & \multicolumn{4}{c}{AP~(\%)} & \multirow{2}{*}{mAP~(\%)}\\
        \cmidrule(lr){2-5}
                & \centering drink & \centering face & \centering phone & \centering smoke & \\
        \midrule
        YOLOv7-X (Baseline) \cite{yolov7} & 93.71 & 93.01 & 81.16 & 86.82 & 88.68\\
        YOLOv3 \cite{yolov3} & 89.13 & 86.23 & 65.32 & 72.62 & 78.32\\
        YOLOv4 \cite{yolov4} & 88.38 & 76.33 & 62.55 & 62.76 & 72.50\\
        SSD \cite{ssd} & 76.91 & 86.16 & 62.52 & 58.50 & 71.02\\
        Faster R-CNN \cite{faster} & 85.34 & 90.59 & 70.89 & 68.21 & 78.76\\
        YOLOv8-L \cite{yolov8} & \underline{95.54} & \textbf{96.76} & \underline{82.20} & \underline{87.61} & \underline{90.53}\\
        YOLOv7-Tiny\cite{yolov7} & 87.29 & 89.29 & 71.44 & 74.06 & 80.52\\
        
        \rowcolor{cvprblue!20}
        \textbf{TACR-YOLO (ours)} & \textbf{95.97} & \underline{95.64} & \textbf{86.68} & \textbf{89.38} & \textbf{91.92}\\
        \hline
   \end{tabular}}
\label{comparsion}
\end{table}

\subsection{Performance Comparisons}

To provide a thorough evaluation of TACR-YOLO, we compare its performance against several leading object detection algorithms using the PABD dataset, as presented in Table~\ref{comparsion}. TACR-YOLO achieves significantly higher AP scores and mAP than YOLOv3~\cite{yolov3}, YOLOv4~\cite{yolov4}, SSD~\cite{ssd}, Faster R-CNN~\cite{faster}, YOLOv8-L~\cite{yolov8}, and YOLOv7-tiny~\cite{yolov7}. Its dynamic channel modulation enhances feature extraction, small object detection, and semantic representation, while improving both fine-grained feature processing and the receptive field.

The effectiveness of the proposed algorithm design for abnormal behavior detection in natural scenarios is further demonstrated through comparison with mainstream detectors. Partial detection results of YOLOv7-X Baseline and TACR-YOLO are illustrated in Fig.~\ref{detection_comparsion}.

To assess real-time performance, we measured the FPS of TACR-YOLO and other mainstream detectors on the same device using five test samples. As shown in Table~\ref{speed}, TACR-YOLO reaches 24.88 FPS, with almost no degradation in real-time performance compared to YOLOv7-X, while delivering significantly improved detection accuracy.

\begin{figure*}[htbp]
\centering
\includegraphics[width=0.91\linewidth, height=0.23\textheight]{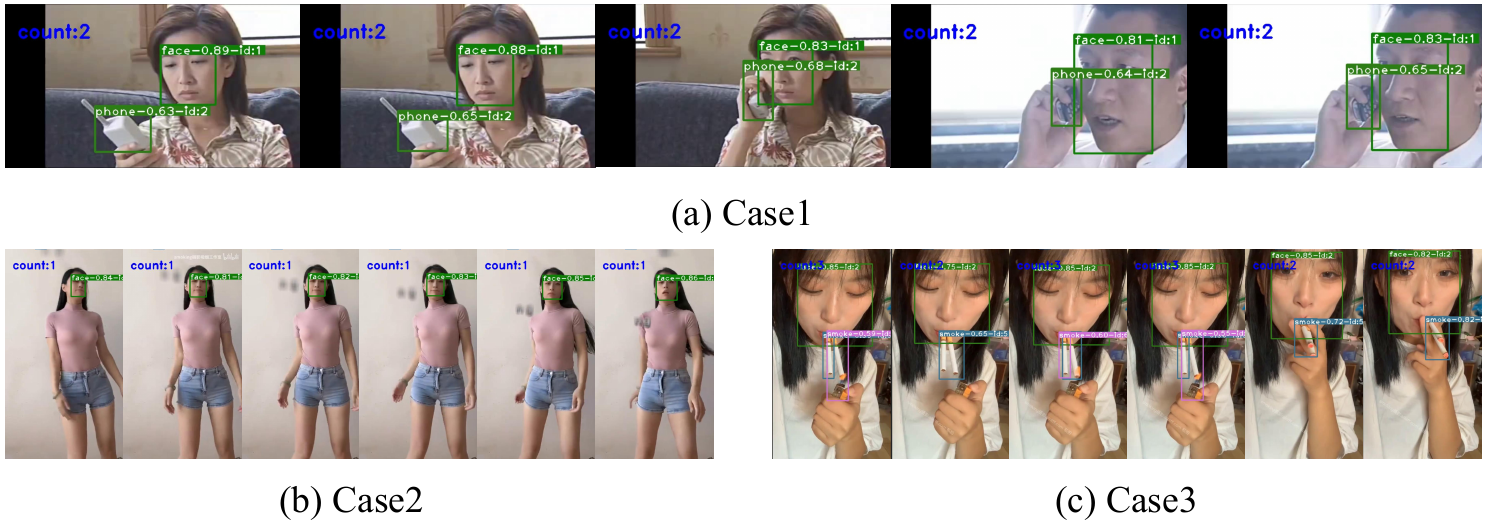}
\vspace{-0.5em}
\caption{The qualitative results of our detection and tracking joint system.}
\label{deepsort-results}
\vspace{-2em}
\end{figure*}

\begin{table}[t!]
\centering
\caption{Inference time comparisons of our TACR-YOLO and other advanced real-time object detection methods on one NVIDIA V100 GPU.}
\renewcommand{\arraystretch}{1.1} 
\setlength\tabcolsep{3pt} 
\resizebox{\linewidth}{!}{
\begin{tabular}{lccc}
        \toprule
        Methods & Detection Time Per Image (s) $\downarrow$ & FPS $\uparrow$ \\
        \midrule
        YOLOv7-X (Baseline) \cite{yolov7} & 0.039 & 25.65 \\
        YOLOv3 \cite{yolov3} & 0.036 & 27.48 \\
        YOLOv4 \cite{yolov4} & 0.042 & 23.61 \\
        SSD \cite{ssd} & 0.021 & 48.35 \\
        Faster R-CNN \cite{faster} & 0.060 & 16.56 \\
        YOLOv8-L \cite{yolov8} & 0.039 & 25.36 \\
        YOLOv7-Tiny \cite{yolov7} & 0.015 & 63.34 \\
        \rowcolor{cvprblue!20}
        \textbf{TACR-YOLO (ours)} & 0.040 & 24.88 \\
        \hline
    \end{tabular}}
\label{speed}
\end{table}

\subsection{Ablation Studies}
We conduct ablation studies to validate the effectiveness of TACR-YOLO, with the first group using YOLOv7-X as a control. The second group consists of YOLOv7-X combined with K-means Clustering. And the third group, based on the second group, adds the Task-Aware Attention Module. The fourth group added the Coordinate Attention Module based on the third group. The fifth group further introduced the Strengthen Neck Network based on the fourth group. The sixth group used the DIoU Loss as the loss function based on the fifth group, i.e., TACR-YOLO.

\noindent \textbf{The Contribution of the K-means Clustering.}~In the second set of experiments, the object bounding box sizes in the training set are clustered by the K-means Clustering to determine a set of anchor box sizes suitable for the dataset, which improves the model’s performance and adaptability in detecting objects across various scales.

\noindent \textbf{The Contributions of the Task-Aware Attention Module.}~We apply the Task-Aware Attention Module to improve the prediction head based on the second set of experiments, resulting in a significant mAP improvement of 1.22\%. This demonstrates the effectiveness of the proposed module. The Task-Aware Attention Module dynamically selects and activates channels that are better suited for the current task, adjusting channel weights based on the differing requirements of classification—where semantic information is more critical—and bounding box regression—where spatial information is prioritized. Each channel can be partially or fully activated or suppressed depending on task-specific needs, enabling the model to learn task-oriented feature representations more effectively and efficiently.

\noindent \textbf{The Impacts of the Coordinate Attention Module.}~Building upon the third set of experiments, we incorporate the CA Module to guide the features extracted from the backbone, resulting in a 0.82\% improvement in mAP. The CA Module improves performance on multi-scale objects by integrating spatial position information into the channel attention mechanism. By integrating it into the backbone during feature extraction, the model's capacity to learn spatial and semantic features is significantly enhanced, leading to improved robustness in multi-scale object detection.

\noindent \textbf{The Effectiveness of Strengthening Neck Network.}~In the fifth set of experiments, building upon the fourth configuration, we enhance the neck network by replacing the single convolution before the Feature Pyramid Network (FPN) with a three-layer convolutional structure. This modification increases the network’s capacity and depth, expands the receptive field, resulting in a 0.5\% improvement in mAP.

\noindent \textbf{The Contributions of DIOU Loss.}~In the sixth set of experiments, we employed DIoU Loss, which comprehensively considers the Euclidean distance between object and anchor box centers, IoU, and the relative scales of the boxes. Table \ref{ablation} shows that compared to IoU, DIoU improved the model's mAP by 2.5\%. This method stabilizes bounding box regression, avoiding issues such as divergence during training that are common with traditional IoU Loss methods.

We finally implemented five improvements, and finally, the mAP improved from 88.68\% to 91.92\%, with a total increase of 3.24\%, of which the Task-Aware Attention Module and the CA module are the important improvement methods, which respectively improve the mAP by 1.22\% and 0.82\%. Table \ref{ablation} documents the progressive performance changes through module ablation.

\begin{table}[t!]
\centering
\caption{Ablation studies on our improvements for our TACR-YOLO on the proposed PABD dataset.}
\renewcommand{\arraystretch}{1.2} 
\resizebox{\linewidth}{!}{
   \begin{tabular}{lcccccc} 
        \toprule
        \multirow{2}{*}{Methods} & \multicolumn{4}{c}{AP~(\%)} & \multirow{2}{*}{mAP~(\%)}\\
        \cmidrule(lr){2-5}
                & \centering drink & \centering face & \centering phone & \centering smoke & \\
        \midrule
            YOLOv7-X (Baseline) \cite{yolov7} & 93.71 & 93.01 & 81.16 & 86.82 & 88.68\\
            + K-means Clustering & 94.23 & 94.25 & 82.04 & 86.01 & 89.13 \color{red} (+0.45)\\
            + Task-Aware Attention Module & 94.28 & 94.44 & 84.95 & 87.72 & 90.35 \color{red} (+1.22)\\
            + Coordinate Attention Module & 96.39 & 95.07 & 84.12 & 89.12 & 91.17 \color{red} (+0.82)\\
            + Strengthen Neck Network & 96.05 & 95.78 & 85.42 & 89.42 & 91.67 \color{red} (+0.50)\\
            + DIoU Loss & 95.97 & 95.64 & 86.68 & 89.38 & 91.92 \color{red} (+0.25)\\
            \bottomrule
        \end{tabular}}
\label{ablation}
\end{table}

\subsection{Downstream Applications}
To further demonstrate the effectiveness of the TACR-YOLO, we deploy and apply our TACR-YOLO, combining it with Deep-SORT \cite{deepsort} to realize an efficient and real-time multi-object tracking system. Deep-SORT \cite{deepsort} is a tracking-by-detection algorithm that combines a convolutional neural network with Kalman filtering to accurately track multiple objects in video streams. As illustrated in Fig.~\ref{deepsort-results}, our coupled system can globally achieve impressive detection and tracking performance.

\section{Conclusion and Discussions}
This paper proposes TACR-YOLO for detecting abnormal human behaviors and introduces a new dataset named PABD, which comprises 8,529 images spanning four categories and covers a wider range of scenarios. By integrating the Coordinate Attention Module, our model effectively captures both spatial and semantic information of multi-scale objects. Additionally, the Task-aware Attention Module dynamically selects and activates channels most relevant to the current task, thereby enhancing the performance of the prediction head. The performance of TACR-YOLO on the PABD dataset demonstrates its robustness and effectiveness, meeting the practical demands for accurate and reliable detection of anomalous behaviors in real-world scenarios.

In the future development, we mainly focus on further pruning and optimizing the overall the model while maintaining the impressive performance.


\bibliographystyle{IEEEbib}
\bibliography{ijcnn2025references}

\end{document}